%% file: travelgan.tex
\documentclass[10pt,twocolumn,letterpaper]{article}

\usepackage{cvpr}
\usepackage{times}
\usepackage{epsfig}
\usepackage{graphicx}
\usepackage{amsmath}
\usepackage{amssymb}
\usepackage{color}
\usepackage{subcaption}
\usepackage{float}
\usepackage[inline]{enumitem}

\newcommand{\beginsupplement}{%
        \setcounter{table}{0}
        \renewcommand{\thetable}{S\arabic{table}}%
        \setcounter{figure}{0}
        \renewcommand{\thefigure}{S\arabic{figure}}%
     }
     
\usepackage[pagebackref=true,breaklinks=true,letterpaper=true,colorlinks,bookmarks=false]{hyperref}

\cvprfinalcopy 

\def\cvprPaperID{4331} 

\ifcvprfinal\fi

\usepackage[utf8]{inputenc}
\begin{document}
\title{TraVeLGAN: Image-to-image Translation by Transformation Vector Learning}

\author{Matthew Amodio\\
Yale University\\
{\tt\small matthew.amodio@yale.edu}
\and
Smita Krishnaswamy\\
Yale University\\
{\tt\small smita.krishnaswamy@yale.edu}
}

\maketitle

\begin{abstract}
\input{abstract.tex}
\end{abstract}

\section{Introduction}
\input{introduction.tex}

\section{Model}

\input{model.tex}

\section{Experiments}
\input{experiments.tex}


\section{Discussion}
\input{discussion.tex}

{\small
\bibliographystyle{ieee}
\bibliography{bibliography}
}

\pagebreak
\newpage
\clearpage

\beginsupplement
\input{supplemental.tex}

\end{document}

%% file: abstract.tex
Interest in image-to-image translation has grown substantially in recent years with the success of unsupervised models based on the cycle-consistency assumption. The achievements of these models have been limited to a particular subset of domains where this assumption yields good results, namely homogeneous domains that are characterized by style or texture differences. We tackle the challenging problem of image-to-image translation where the domains are defined by high-level shapes and contexts, as well as including significant clutter and heterogeneity. For this purpose, we introduce a novel GAN based on preserving intra-domain vector transformations in a latent space learned by a siamese network. The traditional GAN system introduced a discriminator network to guide the generator into generating images in the target domain. To this two-network system we add a third: a siamese network that guides the generator so that each original image shares semantics with its generated version. With this new three-network system, we no longer need to constrain the generators with the ubiquitous cycle-consistency restraint. As a result, the generators can learn mappings between more complex domains that differ from each other by large differences - not just style or texture).

%% file: introduction.tex
\begin{figure}
\centering
\vspace{-8pt}
\includegraphics[width=.7\linewidth]{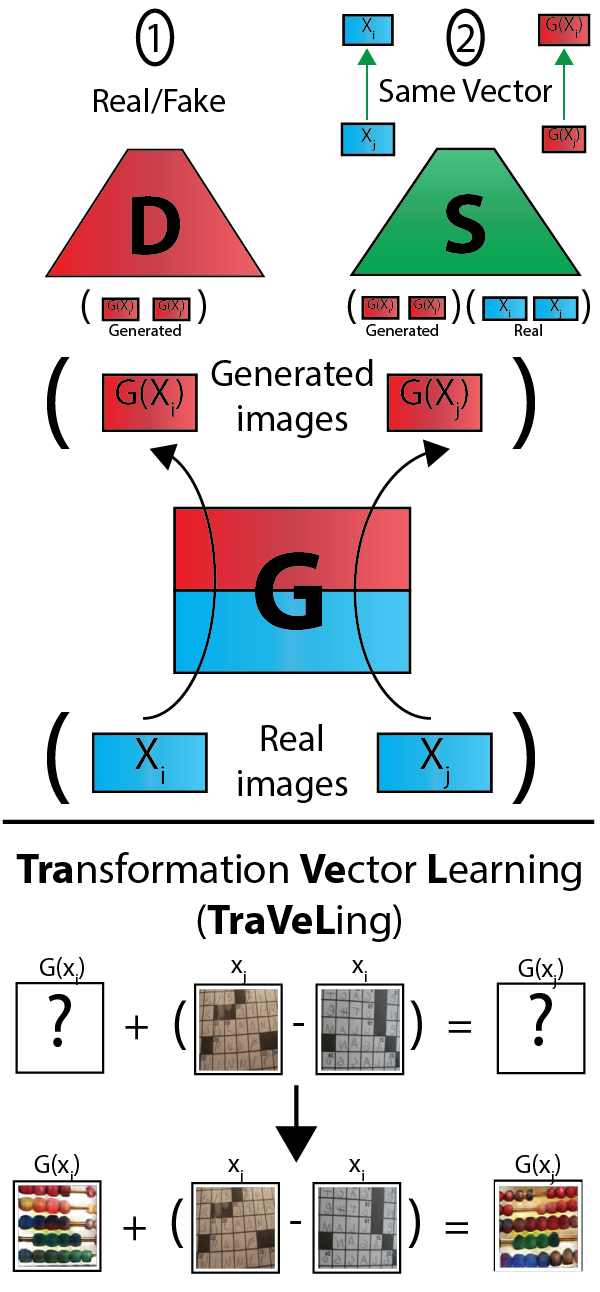}
\vspace{-8pt}
\caption{\small The TraVeLGAN architecture, which adds a siamese network $S$ to the traditional generator $G$ and discriminator $D$ and trains to preserve vector arithmetic between points in the latent space of $S$.}
\label{fig:architecture}
\vspace{-8pt}
\end{figure}

\begin{figure*}
\centering
\vspace{-8pt}
\includegraphics[width=.75\linewidth]{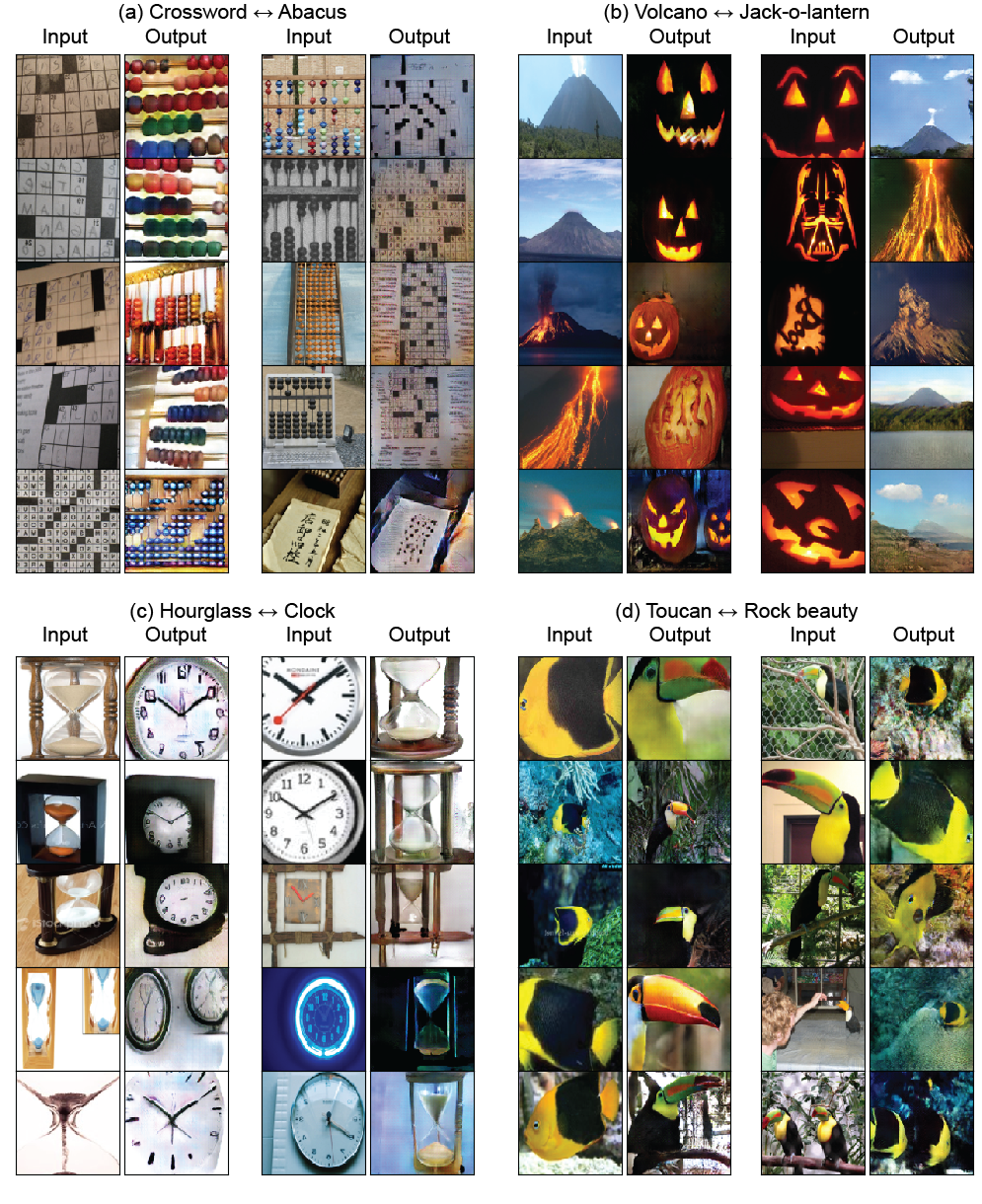}
\vspace{-8pt}
\caption{\small Examples of TraVeLGAN generated output on Imagenet domains that are too different and diverse for cycle-consistent GANs to map between. The TraVeLGAN successfully generates images that are both fully realistic in the output domain (shape of object, color, background) and have preserved semantics learned by the siamese network.}
\label{fig:examples_imagenet}
\vspace{-8pt}
\end{figure*}

Learning to translate an image from one domain to another has been a much studied task in recent years~\cite{yao2015semi,huang2017arbitrary,hoffman2013efficient,zhang2013style,gatys2016image}. The task is intuitively defined when we have paired examples of an image in each domain, but unfortunately these are not available in many interesting cases. Enthusiasm has grown as the field has moved towards unsupervised methods that match the distributions of the two domains with generative adversarial networks (GANs)~\cite{isola2017image,dumoulin2016adversarially,russo2017source,taigman2016unsupervised,liu2016coupled}. However, there are infinitely many mappings between the two domains~\cite{lindvall2002lectures}, and there is no guarantee that an individual image in one domain will share any characteristics with its representation in the other domain after mapping.

Other methods have addressed this non-identifiability problem by regularizing the family of generators in various ways, including employing cross-domain weight-coupling in some layers~\cite{liu2016coupled} and decoding from a shared embedding space~\cite{liu2017unsupervised}. By far the most common regularization, first introduced by the CycleGAN and the DiscoGAN, has been forcing the generators to be each other's inverse, known as the cycle-consistency property~\cite{hoffman2017cycada,zhu2017unpaired,kim2017learning,royer2017xgan,lu2017conditional,almahairi2018augmented,choi2017stargan,anoosheh2017combogan,yi2017dualgan}. Recent findings have shown that being able to invert a mapping at the entire dataset level does not necessarily lead to the generation of related real-generated image pairs~\cite{li2017alice,amodio2018magan,dumoulin2016adversarially}. 

Not only do these dataset-level regularizations on the generator not provide individual image-level matching, but also by restricting the generator, they prevent us from learning mappings that may be necessary for some domains. Previous work continues to pile up regularization after regularization, adding restrictions on top of the generators needing to be inverses of each other. These include forcing the generator to be close to the identity function~\cite{zhu2017unpaired}, matching population statistics of discriminator activations~\cite{kim2017learning}, weight sharing~\cite{liu2016coupled}, penalizing distances in the latent space~\cite{royer2017xgan}, perceptual loss on a previously trained model~\cite{liu2017unsupervised}, or more commonly, multiple of these.

Instead of searching for yet another regularization on the generator itself, we introduce an entirely novel approach to the task of unsupervised domain mapping: the \textbf{Tra}nsformation \textbf{Ve}ctor \textbf{L}earning GAN (TraVeLGAN).

The TraVeLGAN uses a third network, a siamese network, in addition to the generator and discriminator to produce a latent space of the data to capture high-level semantics characterizing the domains. This space guides the generator during training, by forcing the generator to preserve vector arithmetic between points in this space. The vector that transforms one image to another in the original domain must be the same vector that transforms the generated version of that image into the generated version of the other image. Inspired by word2vec embeddings~\cite{goldberg2014word2vec} in the natural language space, if we need to transform one original image into another original image by moving a foreground object from the top-left corner to the bottom-right corner, then the generator must generate two points in the target domain separated by the same transformation vector. 

In word2vec, semantic vector transformations are a \textit{property} of learning a latent space from known word contexts. In TraVeLGAN, we \textit{train} to produce these vectors while learning the space.

Domain mapping consists of two aspects: (a) transfer the given image to the other domain and (b) make the translated image similar to the original image in some way. Previous work has achieved (a) with a separate adversarial discriminator network, but attempted (b) by just restricting the class of generator functions. We propose the natural extension to instead achieve (b) with a separate network, too.

The TraVeLGAN differs from previous work in several substantial ways.
\begin{enumerate}

    \vspace{-6pt}
    \item It completely eliminates the need for training on cycle-consistency or coupling generator weights or otherwise restricting the generator architecture in any way.
    
    \vspace{-6pt}
    \item  It introduces a separate network whose \textit{output} space is used to score similarity between original and generated images. Other work has used a shared latent \textit{embedding} space, but differs in two essential ways: (a) their representations are forced to overlap (instead of preserving vector arithmetic) and (b) the decoder must be able to decode out of the embedding space in an autoencoder fashion~\cite{liu2017unsupervised,royer2017xgan} (\cite{liu2017unsupervised} shows this is in fact equivalent to the cycle consistency constraint.
    
    \vspace{-6pt}
    \item It is entirely parameterized by neural networks: nowhere are Euclidean distances between images assumed to be meaningful by using mean-squared error.
    
    \vspace{-6pt}
    \item It adds interpetability to the unsupervised domain transfer task through its latent space, which explains what aspects of any particular image were used to generate its paired image. 
    
    \vspace{-6pt}
\end{enumerate}
As a consequence of these differences, the TraVeLGAN is better able to handle mappings between complex, heterogeneous domains that require significant and diverse shape changing.

\begin{figure*}
\centering
\vspace{-8pt}
\includegraphics[width=\linewidth]{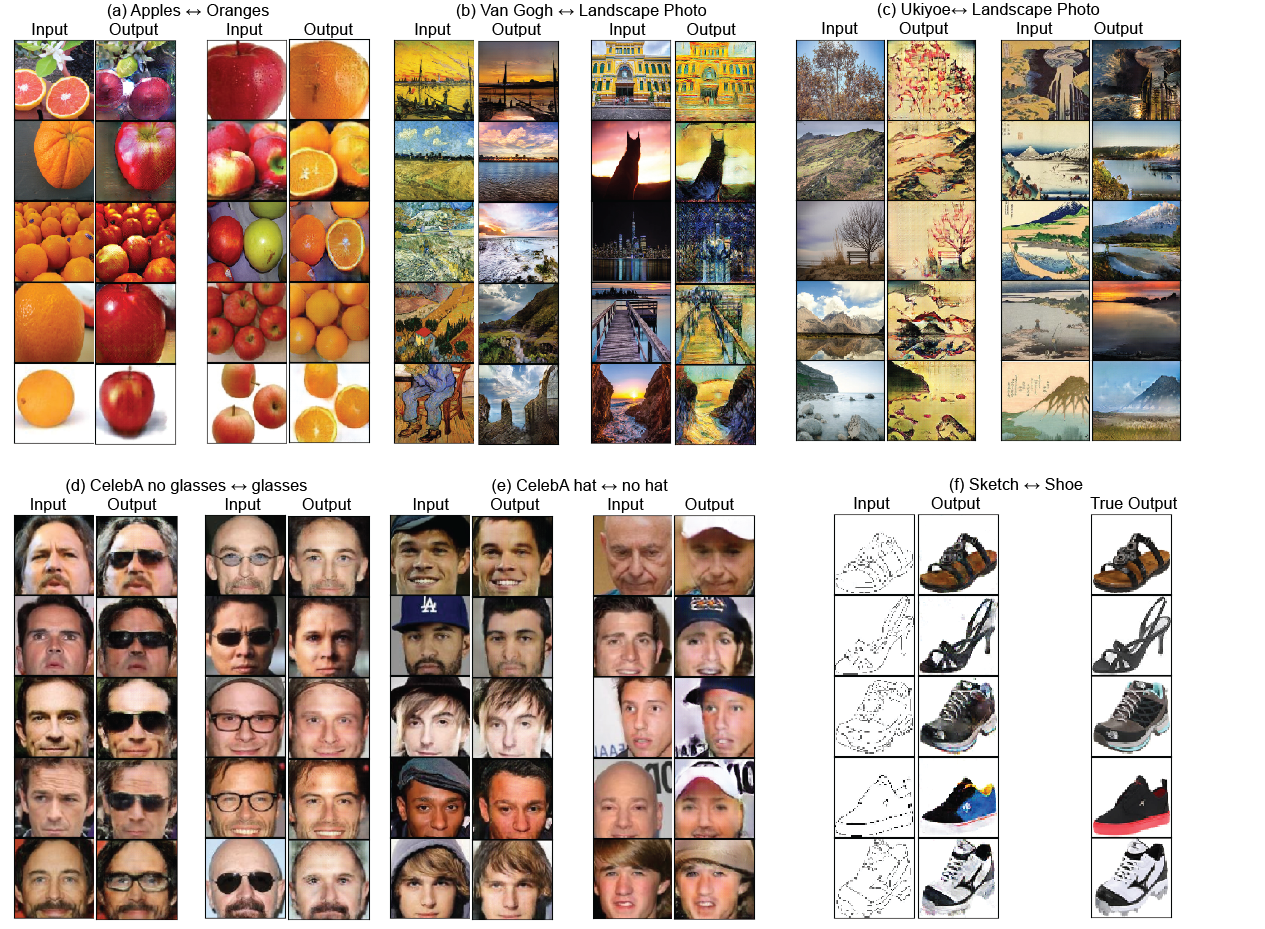}
\vspace{-8pt}
\caption{\small Examples of TraVeLGAN generated output on traditional datasets for unsupervised domain transfer with cycle-consistent GANs. Little change to the original image is necessary in these problems, and TraVeLGAN generates the expected, minimally changed image in the other domain.}
\label{fig:examples_stddata}
\vspace{-8pt}
\end{figure*}


By avoiding direct regularization of the generators, the TraVeLGAN also avoids problems that these regularizations cause. For example, cycle-consistency can unnecessarily prefer an easily invertible function to a possibly more coherent one that is slightly harder to invert (or preventing us from mapping to a domain if the inverse is hard to learn). Not only must each generator learn invertible mappings, but it further requires that the two invertible mappings be each other's inverses. Furthermore, cycle-consistency is enforced with a pixel-wise MSE between the original and reconstructed image: other work has identified the problems caused by using pixelwise MSE, such as the tendency to bias towards the mean images~\cite{bousmalis2017unsupervised}.

Our approach bears a resemblance to that of the DistanceGAN~\cite{benaim2017one}, which preserves pairwise distances between images after mapping. However, they calculate distance directly on the pixel space, while also not preserving any notion of directionality in the space between images. In this paper, we demonstrate the importance of not performing this arithmetic in the pixel space.

Many of these previous attempts have been developed specifically for the task of style transfer, explicitly assuming the domains are characterized by low-level pixel differences (color, resolution, lines) as opposed to high-level semantic differences (shapes and types of specific objects, composition)~\cite{bousmalis2017unsupervised,yi2017dualgan,gatys2016image}. We demonstrate that these models do not perform as well at the latter case, while the TraVeLGAN does.

%% file: model.tex
We denote two data domains $X$ and $Y$, consisting of finite (unpaired) training points $\{x_i\}_{i=1}^{N_x} \in X$ and $\{y_i\}_{i=1}^{N_y} \in Y$, respectively. We seek to learn two mappings, $G_{XY}: X \rightarrow Y$ and $G_{YX}: Y \rightarrow X$, that map between the domains. Moreover, we want the generators to do more than just mimic the domains at an aggregate level. We want there to be a meaningful and identifiable relationship between the two representations of each point. We claim that this task of unsupervised domain mapping consists of two components: \textbf{domain membership} and \textbf{individuality}. Without loss of generality, we define these terms with respect to $G_{XY}$ here, with $G_{YX}$ being the same everywhere but with opposite domains.

\paragraph{Domain membership} The generator should output points in the target domain, i.e. $G_{XY}(X) \in Y$. To enforce this, we use the standard GAN framework of having a discriminator $D_Y$ that tries to distinguish the generator's synthetic output from real samples in $Y$. This yields the typical adversarial loss term $L_{adv}$:
\begin{gather}
L_{adv} = E_X \left [ D_Y(G_{XY}(X)) \right ] \nonumber
\end{gather}

\paragraph{Individuality} In addition, our task has a further requirement than just two different points in $X$ each looking like they belong to $Y$. Given $x_i, x_j \in X, i \ne j$, we want there to be some relationship between $x_i$ and $G_{XY}(x_i)$ that justifies why $G_{XY}(x_i)$ is the representation in domain $Y$ for $x_i$ and not for $x_j$. Without this requirement, the generator could satisfy its objective by ignoring anything substantive about its input and producing arbitrary members of the other domain.

While other methods try to address this by regularizing $G_{XY}$ (by forcing it to be close to the identity or to be inverted by $G_{YX}$), this limits the ability to map between domains that are too different. So instead of enforcing similarity between the point $x_i$ and the point $G_{XY}(x_i)$ directly in this way, we do so implicitly by matching the relationship between the $x_i$'s and the relationship between the corresponding $G_{XY}(x_i)$'s.

We introduce the notion of a \textit{transformation vector} between two points. In previous natural language processing applications~\cite{goldberg2014word2vec}, there is a space where the vector that would transform the word \textit{man} to the word \textit{woman} is similar to the vector that would transform \textit{king} to \textit{queen}. In our applications, rather than changing the gender of the word, the transformation vector could change the background color, size, or shape of an image. The crucial idea, though, is that whatever transformation is necessary to turn one original image into another original image, an analogous transformation must separate the two generated versions of these images. 

Formally, given $x_i, x_j \in X$, define the transformation vector between them $\nu(x_i, x_j) = x_j - x_i$. The generator must learn a mapping such that $\nu(x_i, x_j) = \nu(G_{XY}(x_i), G_{XY}(x_j))$. This is a more powerful property than even preserving distances between points, as it requires the space to be organized such that the directions of the vectors as well as the magnitudes be preserved. This property requires that the vector that takes $x_i$ to $x_j$, be the same vector that takes $G_{XY}(x_i)$ to $G_{XY}(x_j)$.

As stated so far, this framework would only be able to define simple transformations, as it is looking directly at the input space. By analogy, the word-gender-changing vector transformation does not hold over the original one-hot encodings of the words, but instead holds in some reduced semantic latent space. So we instead redefine the transformation vector to be $\nu(x_i, x_j) = S(x_j) - S(x_i)$, where $S$ is a function that gives a representation of each point in some latent space. Given an $S$ that learns high-level semantic representations of each image, we can use our notion of preserving the transformation vectors to guide generation. We propose to learn such a space with an analogue to the adversarial discriminator $D$ from the traditional GAN framework: a cooperative siamese network $S$.

The goal of $S$ is to map images to some space where the relationship between original images is the same as the relationship between their generated versions in the target domain:
\begin{align}
L_{TraVeL} &= \Sigma \Sigma_{i \ne j} Dist(\nu_{ij}, \nu_{ij}^{\prime}) \nonumber \\
\nu_{ij} &= S(x_i) - S(x_j) \nonumber \\
\nu_{ij}^{\prime} &=  S(G_{XY}(x_i)) - S(G_{XY}(x_j)) \nonumber
\end{align}
where $Dist$ is a distance metric, such as cosine similarity. Note this term involves the parameters of $G$, but $G$ needs this space to learn its generative function in the first place. Thus, these two networks depend on each other to achieve their goals. However, unlike in the case of $G$ and $D$, the goals of $G$ and $S$ are not opposed, but cooperative. They both want $L_{TraVeL}$ to be minimized, but $G$ will not learn a trivial function to satisfy this goal, because it also is trying to fool the discriminator. $S$ could still learn a trivial function (such as always outputting zero), so to avoid this we add one further requirement and make its objective multi-task. It must satisfy the standard siamese margin-based contrastive objective~\cite{melekhov2016siamese,ong2017siamese} $L_{S_{c}}$, that every point is at least $\delta$ away from every other point in the latent space:
\begin{align}
L_{S_c} &= \Sigma \Sigma_{i \ne j} max(0, (\delta - ||\nu_{ij}||_2)) \nonumber
\end{align}
This term incentivizes $S$ to learn a latent space that identifies some differences between images, while $L_{TraVeL}$ incentivizes $S$ to organize it. Thus, the final objective terms of $S$ and $G$ are:
\begin{align}
L_S &=  L_{S_{c}} + L_{TraVeL} \nonumber \\
L_G &= L_{adv} + L_{TraVeL} \nonumber
\end{align}
$G$ and $S$ are cooperative in the sense that each is trying to minimize $L_{TraVeL}$, but each has an additional goal specific to its task as well. We jointly train these networks such that together $G$ learns to generate images that $S$ can look at and map to some space where the relationships between original and generated images are preserved.



\begin{figure}
\centering
\vspace{-8pt}
\includegraphics[width=.8\linewidth]{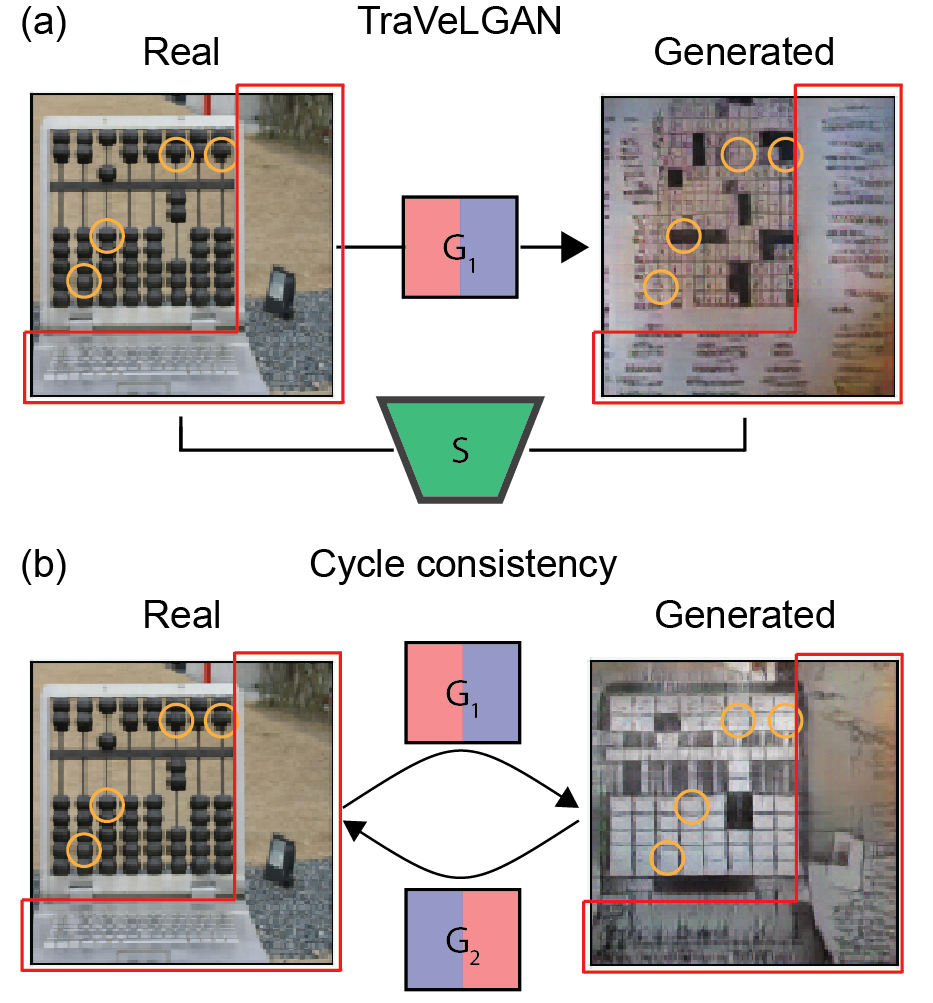}
\vspace{-8pt}
\caption{\small It is hard to learn mappings between domains that are each other's inverse when the domains are asymmetric (e.g. crossword configurations are more complex than abacus configurations). (a) $G_1$ can change the background (red selection) or black beads (orange circles) in hard-to-invert ways. (b) The cycle-consistency assumption forced every black bead to a white crossword square and every blank space to a black crossword square, even though the result is not a realistic crossword pattern. The background is also not fully changed because it could not learn that more complicated inverse function.}
\label{fig:asymmetric}
\vspace{-8pt}
\end{figure}

%% file: experiments.tex
Our experiments are designed around intentionally difficult image-to-image translation tasks. These translations are much harder than style or texture transfer problems, where the domain transformation can be reduced to repeating a common patch transformation on every local patch of every image without higher-level semantic information (e.g. turning a picture into a cartoon)~\cite{royer2017xgan,johnson2016perceptual}. Instead, we choose domains where the differences are higher-level and semantic. For example, when mapping from horses to birds, any given picture of a horse might solely consist of style, texture, and patches that appear in other pictures of real birds (like blue sky, green grass, sharp black outlines, and a brown exterior). Only the higher-level shape and context of the image eventually reveal which domain it belongs to. Additionally, because we use datasets that are designed for classification tasks, the domains contain significant heterogeneity that makes finding commonality within a domain very difficult.

We compare the TraVeLGAN to several previous methods that first regularize the generators by enforcing cycle-consistency and then augment this with further regularizations~\cite{zhu2017unpaired,kim2017learning,amodio2018magan,royer2017xgan,lu2017conditional,almahairi2018augmented,choi2017stargan,anoosheh2017combogan}. Namely, we compare to a GAN with just the cycle-consistency loss (cycle GAN)~\cite{zhu2017unpaired}, with cycle-consistency loss plus the identity regularization (cycle+identity GAN)~\cite{zhu2017unpaired}, with cycle-consistency loss plus a correspondence loss (cycle+corr GAN)~\cite{amodio2018magan}, with cycle-consistency loss plus a feature matching regularization (cycle+featmatch GAN)~\cite{kim2017learning}, and with cycle-consistency loss plus a shared latent space regularization (cycle+latent GAN)~\cite{liu2017unsupervised}. The TraVeLGAN utilizes a U-net architecture with skip connections~\cite{ronneberger2015u} for the generator. The discriminator network is a standard stride-$2$ convolutional classifier network that doubles the number of filters at each layer until the layer is $4x4$ and outputs a single sigmoidal probability. The siamese network is identical except rather than outputting one node like the discriminator it outputs the number of nodes that is the size of the latent space, without any nonlinear activation. For the cycle-consistent GANs we compare to, we optimized the hyperparameters to get the best achievement we could, since our focus is on testing our different loss formulation. This involved trying both Resnet and U-Net architectures for the models from~\cite{zhu2017unpaired}: the U-Net performed much better than the Resnet at these tasks, so we use that here. We also had to choose a value of the cycle-consistent coefficient that largely de-emphasized it in order to get them to change the input image at all ($0.1$). Even so, we were not able to achieve nearly as convincing results with any of the baseline models as with the TraVeLGAN.

\subsection{Similar domains}
The datasets we first consider are traditional cases for unsupervised domain mapping with cycle-consistent networks, where little change is necessary. These are: \vspace{-13pt}
\paragraph{Apples to oranges} The photos of apples and oranges from \cite{zhu2017unpaired} (Figure \ref{fig:examples_stddata}a). The TraVeLGAN successfully changes not only the color of the fruit, but also the shape and texture. The stem is removed from apples, for example, and the insides of oranges aren't just colored red but fully made into apples. In the last row, the TraVeLGAN changes the shape of the orange to become an apple and correspondingly moves its shadow down in the frame to correspond. \vspace{-13pt}
\paragraph{Van Gogh to landscape photo} The portraits by Van Gogh and photos of landscapes, also from \cite{zhu2017unpaired} (Figure \ref{fig:examples_stddata}b). Here the prototypical Van Gogh brush strokes and colors are successfully applied or removed. Notably, in the last row, the portrait of the man is changed to be a photo of a rocky outcrop with the blue clothes of the man changing to blue sky and the chair becoming rocks, rather than becoming a photo-realistic version of that man, which would not belong in the target domain of landscapes. \vspace{-13pt}
\paragraph{Ukiyoe to landscape photo} Another dataset from \cite{zhu2017unpaired}, paintings by Ukiyoe and photos of landscapes (Figure \ref{fig:examples_stddata}c). It is interesting to note that in the generated Ukiyoe images, the TraVeLGAN correctly matches reflections of mountains in the water, adding color to the top of the mountain and the corresponding bottom of the reflection.
\vspace{-13pt}
\paragraph{CelebA glasses} The CelebA dataset filtered for men with and without glasses~\cite{celeba} (Figure \ref{fig:examples_stddata}d). As expected, the TraVeLGAN produces on the minimal change necessary to transfer an image to the other domain, i.e. adding or removing glasses while preserving the other aspects of the image. Since the TraVeLGAN learns a semantic, rather than pixel-wise, information preserving penalty, in some cases aspects not related to the domain are also changed (like hair color or background). In each case, the resulting image is still a convincingly realistic image in the target domain with a strong similarity to the original, though. \vspace{-13pt}
\paragraph{CelebA hats} The CelebA dataset filtered for men with and without hats~\cite{celeba} (Figure \ref{fig:examples_stddata}e). As before, the TraVeLGAN adds or removes a hat while preserving the other semantics in the image.
\vspace{-13pt}
\paragraph{Sketch to shoe} Images of shoes along with their sketch outlines, from \cite{shoe} (Figure \ref{fig:examples_stddata}f). Because this dataset is paired (though it is still trained unsupervised as always), we are able to quantify the performance of the TraVeLGAN with a heuristic: the pixel-wise mean-squared error (MSE) between the TraVeLGAN's generated output and the true image in the other domain. This can be seen to be a heuristic in the fourth row of Figure \ref{fig:examples_stddata}c, where the blue and black shoe matches the outline of the sketch perfectly, but is not the red and black color that the actual shoe happened to be. However, even as an approximation it provides information. Table \ref{tab:shoe_accuracy} shows the full results, and while the vanilla cycle-consistent network performs the best, the TraVeLGAN is not far off and is better than the others. Given that the TraVeLGAN does not have the strict pixel-wise losses of the other models and that the two domains of this dataset are so similar, it is not surprising that the more flexible TraVeLGAN only performs similarly to the cycle-consistent frameworks.
These scores provide an opportunity to gauge the effect of changing the size of the latent space learned by the siamese network. We see that our empirically chosen default value of $1000$ slightly outperforms a smaller and lower value. This parameter controls the expressive capability of the model, and the scores suggest providing it too small of a space can limit the complexity of the learned transformation and too large of a space can inhibit the training. The scores are all very similar, though, suggesting it is fairly robust to this choice.
\vspace{-13pt}
\paragraph{Quantitative results} Since the two domains in these datasets are so similar, it is reasonble to evaluate each model using structural similarity (SSIM) between the real and generated images in each case. These results are presented in Table~\ref{tab:ssim}. There we can see that the TraVeLGAN performs comparably to the cycle-consistent models. It is expected that the baselines perform well in these cases, as these are the standard applications they were designed to succeed on in the first place; namely, domains that require little change to the original images. Furthermore, it is expected that the TraVeLGAN changes the images slightly more than the models that enforce pixel-wise cycle-consistency. That the TraVeLGAN performs so similarly demonstrates quantitatively that the TraVeLGAN can preserve the main qualities of the image when the domains are similar.

\begin{table}[t]
\centering
\scriptsize
\begin{tabular}{|l|l|l|l|l|l|}
\hline
SSIM & Apple & Van Gogh & Ukiyoe & Glasses & Hats \\ \hline \hline
TraVeLGAN &  0.302 & 0.183 & 0.222 & 0.499 & 0.420 \\ \hline
Cycle &  \textbf{0.424} & 0.216 & 0.252 & 0.463 &\textbf{0.437} \\ \hline
Cycle+ident &  0.305 & \textbf{0.327} & \textbf{0.260} & \textbf{0.608} & 0.358 \\ \hline
Cycle+corr &  0.251 & 0.079 & 0.072 & 0.230 & 0.204 \\ \hline
Cycle+featmatch & 0.114 & 0.117 & 0.125 & 0.086 & 0.209 \\ \hline
Cycle+latent & 0.245 & 0.260 & 0.144 & 0.442 & 0.382 \\ \hline
\end{tabular}
\caption{\small Real/generated SSIM on the similar-domains datasets.}
\label{tab:ssim}
\end{table}

\begin{table}[t]
\centering
\scriptsize
\begin{tabular}{|l|l|l|}
\hline
Pixel MSE & Sketches & Shoes \\ \hline \hline
TraVeLGAN & 0.060 & 0.267 \\ \hline
TraVeLGAN ($D_{latent}$=100) & 0.069 & 0.370 \\ \hline
TraVeLGAN ($D_{latent}$=2000) & 0.064 & 0.274 \\ \hline
Cycle & \textbf{0.047} & \textbf{0.148} \\ \hline
Cycle+corr & 0.427 & 0.603 \\ \hline
Cycle+featmatch & 0.077 & 0.394 \\ \hline
Cycle+latent & 0.072 & 0.434 \\ \hline
\end{tabular}
\caption{\small Per-pixel MSE on the shoes-to-sketch dataset.}
\label{tab:shoe_accuracy}
\end{table}

\subsection{Imagenet: diverse domains}
The previous datasets considered domains that were very similar to each other. Next, we map between two domains that are not only very different from each other, but from classification datasets where the object characterizing the domain is sometimes only partially in the frame, has many different possible appearances, or have substantial clutter around it. In this most difficult task, we present arbitrary chooses two classes from the Imagenet~\cite{deng2009imagenet} dataset. These images are much higher-resolution (all images are rescaled to $128$x$128$), making it easier to learn a transfer that only needs local image patches (like style/texture transfer) than entire-image solutions like TraVeLGAN's high-level semantic mappings.

We chose classes arbitrarily because we seek a framework that is flexible enough to make translations between any domains, even when those classes are very different and arbitrarily picked (as opposed to specific domains contrived to satisfy particular assumptions). The pairs are:
\begin{enumerate*}
\item abacus and crossword (Figure \ref{fig:examples_imagenet}a)
\item volcano and jack-o-lantern (Figure \ref{fig:examples_imagenet}b)
\item clock and hourglass (Figure \ref{fig:examples_imagenet}c)
\item toucan and rock beauty (Figure \ref{fig:examples_imagenet}d).
\end{enumerate*}

\begin{figure}
\centering
\vspace{-8pt}
\includegraphics[width=\linewidth]{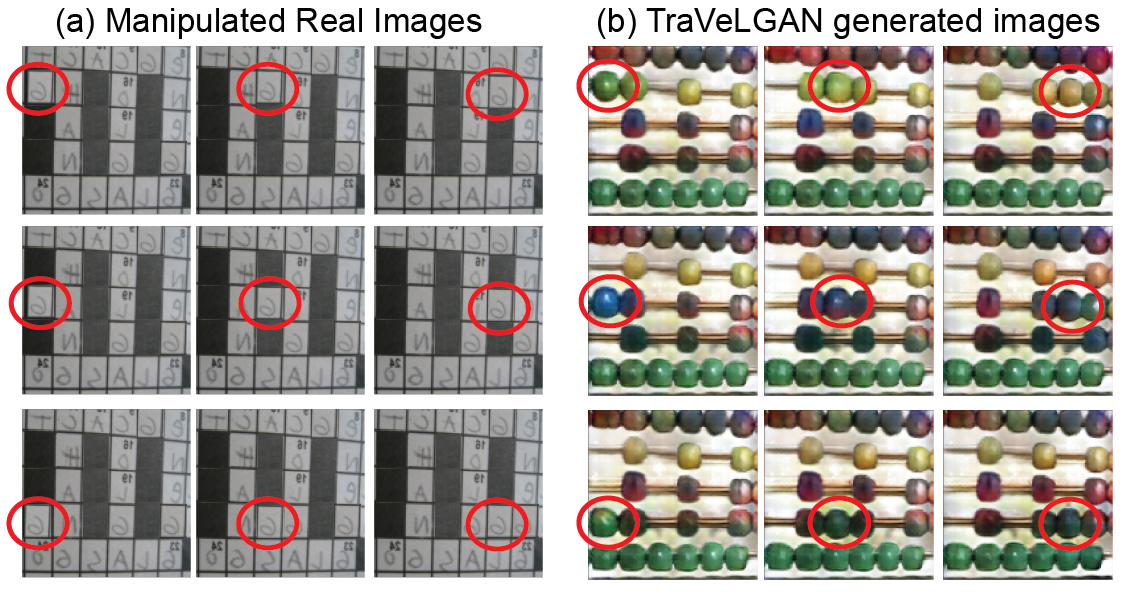}
\vspace{-8pt}
\caption{\small (a) A real crossword image artificially manipulated to move a white square around the frame. (b) The TraVeLGAN, which has not seen any of these images during training, has learned a semantic mapping between the domains that moves an abacus bead appropriately with the crossword square.}
\label{fig:mutations}
\vspace{-8pt}
\end{figure}

\begin{figure}
\centering
\vspace{-8pt}
\includegraphics[width=\linewidth]{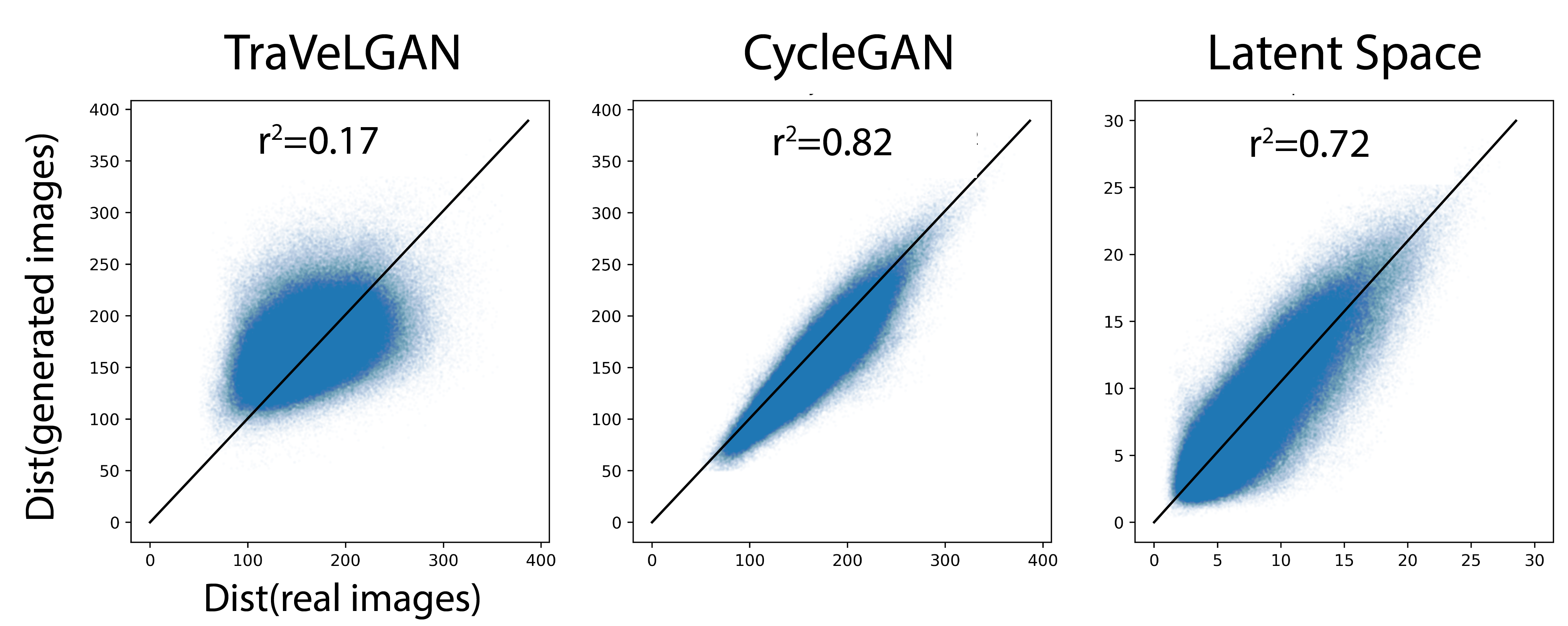}
\vspace{-8pt}
\caption{\small The CycleGAN generates images such that pairwise L2-distances in \textit{\textbf{pixel~space}} are strongly preserved. The TraVeLGAN generated images are virtually uncorrelated in pixel space, but the siamese network learns a \textit{\textbf{latent~space}} where pairwise distances are preserved.}
\label{fig:correlated_distances}
\vspace{-8pt}
\end{figure}

\vspace{-12pt}
\paragraph{Asymmetric domains}
Learning to map between the domains of abacus and crossword showcase a standard property of arbitrary domain mapping: the amount and nature of variability in one domain is larger than in the other. In Figure \ref{fig:asymmetric}, we see that the TraVeLGAN learned a semantic mapping from an abacus to a crossword by turning the beads of an abacus into the white squares in a crossword and turning the string in the abacus to the black squares. However, in an abacus, the beads can be aligned in any shape, while in crosswords only specific grids are feasible. To turn the abacus in Figure \ref{fig:asymmetric} (which has huge blocks of beads that would make for a very difficult crossword indeed!) into a realistic crossword, the TraVeLGAN must make some beads into black squares and others into white squares. The cycle-consistency loss fights this one-to-many mapping because it would be hard for the other generator, which is forced to also be the inverse of this generator, to learn the inverse many-to-one function. So instead, it learns a precise, rigid bead-to-white-square and string-to-black-square mapping at the expense of making a realistic crossword (Figure \ref{fig:asymmetric}b). Even though the background is an unimportant part of the image semantically, it must recover all of the exact pixel-wise values after cycling. We note that the TraVeLGAN automatically relaxed the one-to-one relationship of beads to crossword squares to create realistic crosswords. On the other hand, any real crossword configuration is a plausible abacus configuration. In the next section, we show that the TraVeLGAN also automatically discovered this mapping can be one-to-one in white-squares-to-beads, and preserves this systematically.

\vspace{-12pt}
\paragraph{Manipulated images study}
Next we examine the degree to which the TraVeLGAN has learned a meaningful semantic mapping between domains. Since the Imagenet classes are so cluttered and heterogeneous and lack repetition in the form of two very similar images, we create similar images with a manipulation study. We have taken one of the real images in the crossword domain, and using standard photo-editing software, we have created systematically related images. With these systematically related images, we can test to see whether the TraVeLGAN's mapping preserves the semantics in the abacus domain in a systematic way, too.

In Figure \ref{fig:mutations}, we started with a crossword and created a regular three-by-three grid of black squares by editing an image from Figure \ref{fig:abacus2crossword}. Then, systematically, we move a white square around the grid through each of the nine positions. In each case, the TraVeLGAN generates an abacus with a bead moving around the grid appropriately. Remarkably, it even colors the bead to fit with the neighboring beads, which differ throughout the grid. Given that none of the specific nine images in Figure \ref{fig:mutations} were seen in training, the TraVeLGAN has clearly learned the semantics of the mapping rather than memorizing a specific point.

\begin{table}[t]
\centering
\scriptsize
\begin{tabular}{|l|l|l|l|l|}
\hline
FID score           & (a)              & (b)                & (c)             & (d)          \\ \hline \hline
TraVeLGAN           & \textbf{1.026}   & \textbf{0.032}     & \textbf{0.698}  & \textbf{0.206}\\ \hline
Cycle               & 1.350            & 1.281               & 1.018           & 0.381       \\ \hline
Cycle+identity      & 1.535            & 0.917              & 1.297           & 1.067        \\ \hline
Cycle+corr          & 1.519            & 0.527              & 0.727           & 0.638        \\ \hline
Cycle+featmatch     & 1.357            & 1.331              & 1.084           & 0.869        \\ \hline
Cycle+latent        & 1.221            & 0.485              & 1.104           & 0.543        \\ \hline
\end{tabular}
\caption{\small FID scores for each of the models on each of the Imagenet datasets. Column labels correspond to Figure \ref{fig:examples_imagenet}.}
\label{tab:fid}
\end{table}

\begin{table}[t]
\scriptsize
\centering
\begin{tabular}{|l|l|l|l|l|}
\hline
Discriminator score & (a)              & (b)                & (c)             & (d)                  \\ \hline \hline
TraVeLGAN           & \textbf{0.035}   & \textbf{0.206}     &  \textbf{0.074} & \textbf{0.145} \\ \hline
Cycle               & 0.014            & 0.008              & 0.033           & 0.008                \\ \hline
Cycle+identity      & 0.011            & 0.044              & 0.040           & 0.064                \\ \hline
Cycle+corr          & 0.009            & 0.191              & 0.026           & 0.001                \\ \hline
Cycle+featmatch     & 0.002            & 0.029              & 0.066           & 0.014                \\ \hline
Cycle+latent        & 0.009            & 0.069              & 0.047           & 0.039                \\ \hline       
\end{tabular}
\caption{\small Discriminator scores for each of the models on each of the Imagenet datasets. Column labels correspond to Figure \ref{fig:examples_imagenet}.}
\label{tab:discriminator_scores}
\end{table}

\paragraph{Pairwise distance preservation}
The DistanceGAN \cite{benaim2017one} has shown that approximately maintaining pairwise distances between images in \textit{\textbf{pixel space}} achieves similar success to the cycle-consistent GANs. In fact, they show that cycle-consistent GANs produce images that preserve the pixel pairwise distance between images with extremely highly correlation. On the toucan to rock beauty dataset, we observe the same phenomenon ($r^2=0.82$ in Figure~\ref{fig:correlated_distances}). While this produced plausible images in some cases, maintaining pixel-wise distance between images could not generate realistic toucans or rock beauties. The TraVeLGAN pairwise distances are virtually uncorrelated in pixel space ($r^2=0.17$). However, we understand the role of the siamese network when we look at the pairwise distances between real images in \textit{\textbf{latent space}} and the corresponding pairwise distances between generated images in \textit{\textbf{latent space}}. There we see a similar correlation ($r^2=0.72$). In other words, the TraVeLGAN simultaneously learns a mapping with a neural network to a space where distances can be meaningfully preserved while using that mapping to guide it in generating realistic images.

\paragraph{Quantitative results}
Lastly, we add quantitative evidence to the qualitative evidence already presented that the TraVeLGAN outperforms existing models when the domains are very different. While we used the SSIM and pixel-wise MSE in the previous sections to evaluate success, neither heuristic is appropriate for these datasets. The goal in these mappings is \textit{not} to leave the image unchanged and as similar to the original as possible, it is to fully change the image into the other domain. Thus, we apply use two different metrics to evaluate the models quantitatively on these Imagenet datasets.

In general, quantifying GAN quality is a hard task~\cite{barratt2018note}. Moreover, here we are specifically interested in how well a generated image is paired or corresponding to the original image, point-by-point. To the best of our knowledge, there is no current way to measure this quantitatively for arbitrary domains, so we have pursued the qualitative evaluations in the previous sections. However, in addition to those qualitative evaluation of the correspondence aspect, we at least quantify how well the generated images resemble the target domain, at a population level, with heuristic scores designed to measure this under certain assumptions. The first, the Fréchet Inception Distance (FID score)~\cite{fid} is an improved version of the Inception Score (whose flaws were well articulated in \cite{barratt2018note}) which compares the real and generated images in a layer of a pre-trained Inception network (Table~\ref{tab:fid}). The second, the discriminator score, trains a discriminator from scratch, independent of the one used during training, to try to distinguish between real and generated examples (Table~\ref{tab:discriminator_scores}). The TraVeLGAN achieved better scores than any of the baseline models with both metrics and across all datasets.

%% file: discussion.tex
In recent years, unsupervised domain mapping has been dominated by approaches building off of the cycle-consistency assumption and framework. We have identified that some cluttered, heterogeneous, asymmetric domains cannot be successfully mapped between by generators trained on this cycle-consistency approach. Further improving the flexibility of domain mapping models may need to proceed without the cycle-consistent assumption, as we have done here.

%% file: supplemental.tex






\clearpage

\paragraph{Quantitative results}
Quantitative results are summarized by the FID score (Table \ref{tab:fid}) and the discriminator score (Table \ref{tab:discriminator_scores}). We note that these scores were both designed to evaluate models that attempt to generate the full diversity of the Imagenet dataset, while in our case we only map to a single class.

The Fréchet Inception Distance (FID score)~\cite{fid} calculates the Fréchet distance between Gaussian models of the output of a the pre-trained Inception network~\cite{szegedy2015going} on real and generated images, respectively. Lower distances indicate better performance. The results are the mean of the scores from each direction.

The discriminator score is calculated by training a new discriminator, distinct from the one used during training, to distinguish between real and generated images in a domain. A score of zero means the discriminator was certain every generated image was fake, while higher scores indicate the generated images looked more like real images. As in the FID, the results are the mean of the scores from each direction.

\paragraph{Optimization and training parameters}
Optimization was performed with the adam~\cite{kingma2014adam} optimizer with a learning rate of $0.0002$, $\beta_1=0.5$, $\beta_2=0.9$. Gradient descent was alternated between generator and discriminator, with the discriminator receiving real and generated images in distinct batches.

\paragraph{Architecture}
The TraVeLGAN architecture is as follows. Let $d$ denote the size of the image. Let $c_n$ be a standard stride-two convolutional layer with $n$ filters, $t_n$ be a stride-two convolutional transpose layer with kernel size four and $n$ filters, and $f_n$ be a fully connected layer outputting $n$ neurons. The discriminator $D$ has layers until the size of the input is four-by-four, increasing the number of filters by a factor of two each time, up to a maximum of eight times the original number (three layers for CIFAR and five layers for Imagenet). This last layer is then flattened and passed through a fully connected layer. The overall architecture is thus $c_n-c_{2n}-c_{4n}-c_{8n}-c_{8n}-f_1$. The siamese network has the same structure as the discriminator except its latent space has size $1000$, yielding the architecture $c_n-c_{2n}-c_{4n}-c_{8n}-c_{8n}-f_{1000}$. The generator uses the U-Net architecture~\cite{ronneberger2015u} that has skip connections that concatenate the input in the symmetric encoder with the decoder, yielding layers of $c_n-c_{2n}-c_{4n}-c_{4n}-c_{4n}-t_{8n}-t_{8n}-t_{8n}-t_{4n}-t_{2n}-t_{3}$. For the cycle-consistency networks, the architectures of the original implementations were used, with code from \cite{zhu2017unpaired}, \cite{zhu2017unpaired}, \cite{amodio2018magan}, \cite{kim2017learning}, for the cycle, cycle+identity, cycle+corr, and cycle+featmatch, respectively. All activations are leaky rectified linear units with leak of $0.2$, except for the output layers, which use sigmoid for the discriminator, hyperbolic tangent for the generator, and linear for the siamese network. Batch normalization is used for every layer except the first layer of the discriminator. All code was implemented in Tensorflow~\cite{abadi2016tensorflow} on a single NVIDIA Titan X GPU.

\paragraph{CIFAR}
While the CIFAR images~\cite{krizhevsky2009learning} are relatively simple and low-dimensional, it is a deceptively complex task compared to standard domain mapping datasets like CelebA, where they are all centered close-ups of human faces (i.e. their shoulders or hair are in the same pixel locations). The cycle-consistent GANs struggle to identify the characteristic shapes of each domain, instead either only make small changes to the images or focusing on the color tone. The TraVeLGAN, on the other hand, fully transfers images to the target domain. Furthermore, the TraVeLGAN preserves semantics like orientation, background color, body color, and composition in the pair of image (complete comparison results in Figure \ref{fig:bird2horse})

\begin{figure}
\centering
\includegraphics[width=.9\linewidth]{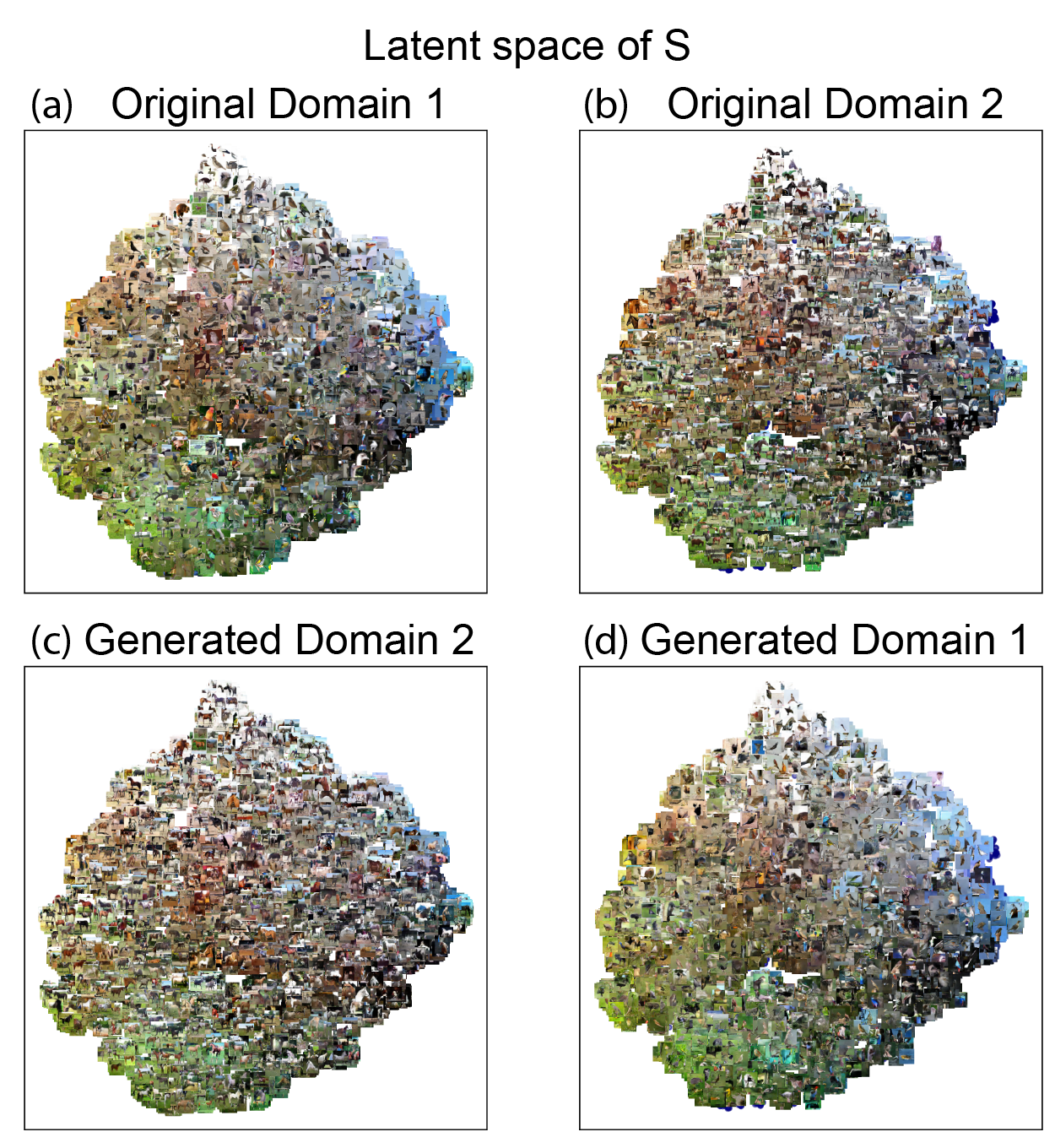}
\caption{Having access to the siamese space output by $S$ provides an interpretability of the TraVeLGAN's domain mapping that other networks lack. PCA visualizations on the CIFAR example indicate $S$ has indeed learned a meaningfully organized space for $G$ to preserve transformation vectors within.}
\label{fig:siamesespace}
\end{figure}

\paragraph{Interpretability}
As the siamese latent space is learned to preserve vector transformations between images, we can look at how that space is organized to tell us what transformation the network learned at a dataset-wide resolution. Figure \ref{fig:siamesespace} shows a PCA visualization of the siamese space of the CIFAR dataset with all of the original domain one (bird) and domain two (horse) images. There we can see that $S$ learned a logical space with obvious structure, where mostly grassy images are in the bottom left, mostly sky images in the top right, and so forth. Furthermore, the layout is analogous between the two domains, verifying that the network automatically learned a notion of similarity between the two domains. We also show every generated image across the whole dataset in this space, where we see that the transformation vectors are not just interpretable for some individual images and not others, but are interpretable across the entire distribution of generated images.

\begin{figure}
\centering
\includegraphics[width=.7\linewidth]{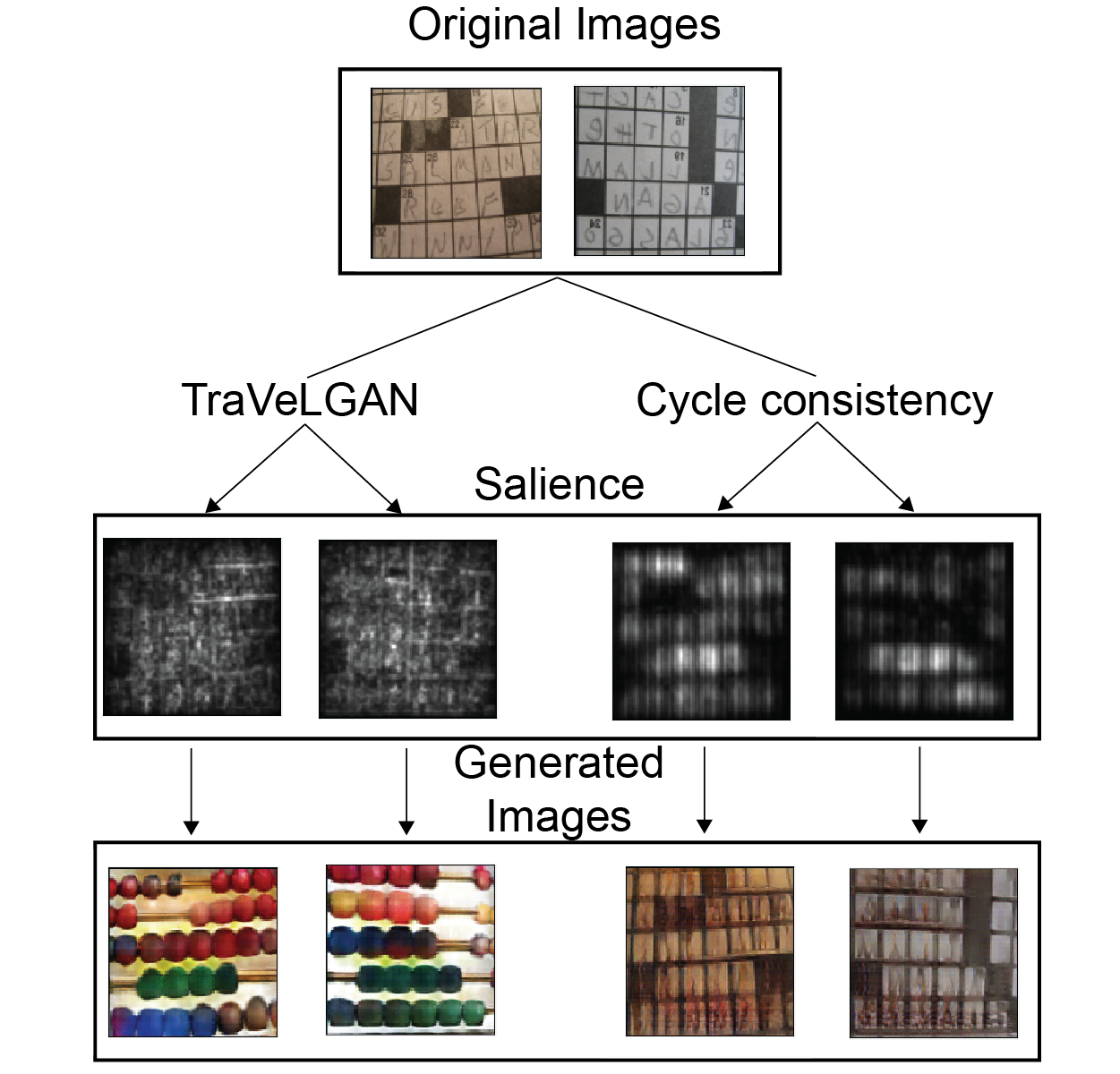}
\caption{A salience analysis exploring how the TraVeLGAN's objective loosens the restriction of cycle-consistency and allows it more flexibility in changing the image during domain transfer. The TraVeL loss requires significantly less memorization of the input pixels, and as a result, more complex transformations can be learned.}
\label{fig:sailences}
\end{figure}

\paragraph{Salience}
We next perform a salience analysis of the TraVeL loss by calculating the magnitude of the gradient at each pixel in the generated image with respect to each pixel in the original image (Figure \ref{fig:sailences}). Since the TraVeL loss, which enforces the similarity aspect of the domain mapping problem, is parameterized by another neural network $S$, the original image contributes to the generated image in a complex, high-level way, and as such the gradients are spread richly over the entire foreground of the image. This allows the generator to make realistic abacus beads, which need to be round and shaded, out of square and uniform pixels in the crossword. By contrast, the cycle-consistency loss requires numerical precision in the pixels, and as such the salience map largely looks like a grayscale version of the real image, with rigid lines and large blocks of homogeneous pixels still visible. This is further evidence that the cycle-consistency loss is preventing the generator from making round beads with colors that vary over the numerical RGB values.
